%% file: main.tex
\documentclass[10pt,twocolumn,letterpaper]{article}

\usepackage[final]{cvpr}      
\makeatletter
\newcommand{\ssymbol}[1]{\ensuremath{^{\@fnsymbol{#1}}}}
\makeatother
\input{preamble}

\definecolor{cvprblue}{rgb}{0.21,0.49,0.74}
\usepackage[pagebackref,breaklinks,colorlinks,allcolors=cvprblue]{hyperref}
\usepackage{url}
\usepackage{hyperref}
\usepackage{algorithm}
\usepackage{algpseudocode}
\usepackage{pifont}
\usepackage{graphicx}
\usepackage{overpic}
\usepackage{booktabs}
\usepackage{makecell}
\usepackage{multirow}
\usepackage{wrapfig}
\usepackage{marvosym}
\usepackage{caption}
\usepackage{colortbl}
\usepackage{xcolor} 
\usepackage{subcaption}
\usepackage{pifont}
\usepackage[table]{xcolor}
\definecolor{aliceblue}{rgb}{0.87, 0.92, 0.96}
\def\paperID{3233} 
\def\confName{CVPR}
\def\confYear{2026}

\title{Unlocking the Forgery Detection Potential of Vanilla MLLMs: \\A Novel Training-Free Pipeline}

\author{Rui Zuo\\
Zhejiang University\\
{\tt\small 22431088@zju.edu.cn}
\and
Qinyue Tong\thanks{Corresponding author.}\\
Zhejiang University\\
{\tt\small qinyuetong@zju.edu.cn}
\and
Zhe-Ming Lu\\
Zhejiang University\\
{\tt\small zheminglu@zju.edu.cn}
\and
Ziqian Lu\\
Zhejiang Sci-Tech University\\
{\tt\small  ziqianlu@zstu.edu.cn}
}

\begin{document}
\input{sec/teaser}
\input{sec/abstract}    
\input{sec/Introduction}

\input{sec/relatedworks}
\input{sec/method}
\input{sec/experiment}
\input{sec/conclusion.tex}

{
    \small
    \bibliographystyle{ieeenat_fullname}
    \bibliography{main}
}


\end{document}

%% file: preamble.tex
\usepackage{xcolor} 

\usepackage{amsmath,amssymb}



%% file: sec/teaser.tex
\twocolumn[{%
\renewcommand\twocolumn[1][]{#1}%
\maketitle
\vspace{-1.3cm}
\begin{center}
    \captionsetup{type=figure}
    \includegraphics[width=0.95\textwidth,keepaspectratio]{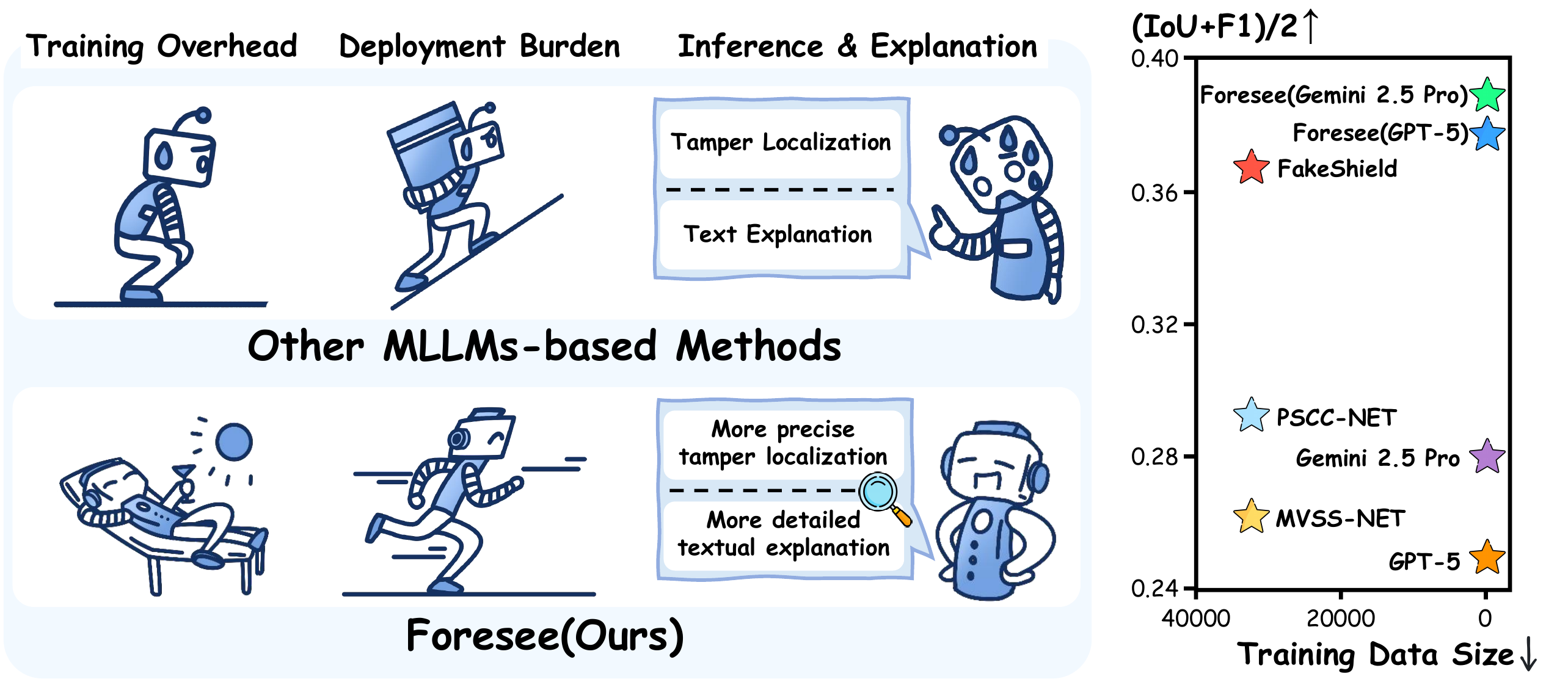} 
    \captionof{figure}{
Comparison of our Foresee pipeline with other state-of-the-art MLLM-based methods and vanilla MLLMs in terms of training overhead, deployment burden, and inference performance. Foresee surpasses existing MLLM-based tamper detection methods by operating without any training, requiring fewer computational and inference resources, providing more precise localization, and delivering richer textual explanations. Notably, vanilla MLLMs achieve a significant improvement in localization performance under our pipeline.
}
    \label{fig:teaser}
\end{center}%
}]

%% file: sec/abstract.tex
\begin{abstract}
With the rapid advancement of artificial intelligence-generated content (AIGC) technologies, including multimodal large language models (MLLMs) and diffusion models, image generation and manipulation have become remarkably effortless. Existing image forgery detection and localization (IFDL) methods often struggle to generalize across diverse datasets and offer limited interpretability. 
Nowadays, MLLMs demonstrate strong generalization potential across diverse vision-language tasks, and some studies introduce this capability to IFDL via large-scale training. However, such approaches cost considerable computational resources, while failing to reveal the inherent generalization potential of vanilla MLLMs to address this problem.
Inspired by this observation, we propose \textbf{Foresee}, \textbf{a training-free MLLM-based pipeline} tailored for image forgery analysis. 
It eliminates the need for additional training and enables a lightweight inference process, while surpassing existing MLLM-based methods in both tamper localization accuracy and the richness of textual explanations. 
Foresee employs a \textbf{type-prior-driven strategy} and utilizes a \textbf{F}lexible \textbf{F}eature \textbf{D}etector (\textbf{FFD}) module to specifically handle copy-move manipulations, thereby effectively unleashing the potential of vanilla MLLMs in the forensic domain. 
Extensive experiments demonstrate that our approach simultaneously achieves superior localization accuracy and provides more comprehensive textual explanations. 
Moreover, Foresee exhibits stronger generalization capability, outperforming existing IFDL methods across various tampering types, including copy-move, splicing, removal, local enhancement, deepfake, and AIGC-based editing. The code will be released in the final version.
\end{abstract}

%% file: sec/Introduction.tex
\section{Introduction}
\label{sec:intro}
With the rapid advancement of artificial intelligence-generated content (AIGC) technologies, including multimodal large language models (MLLMs)~\cite{hurst2024gpt, bai2023qwen, team2023gemini, liu2023visual, tong2025medisee} and diffusion models~\cite{rombach2022high, saharia2022photorealistic}, the generation and manipulation of images have become remarkably effortless. 
In practical scenarios, users simply provide an original image along with their specific requirements to multimodal generative AI models, such as Qwen-VL~\cite{wang2024qwen2} and Sora~\cite{openai_sora_system_card}, which can then perform editing manipulations that previously would have required substantial manual effort.
This paradigm enables the creation of highly realistic or entirely fictitious images with minimal exertion. 
While such capabilities greatly benefit creative industries, they also introduce significant risks—facilitating the spread of misinformation, enabling identity fraud, and compromising personal privacy. 
Consequently, the development of robust image forgery detection and localization (IFDL) strategies has emerged as an urgent challenge in ensuring social trust and security.

IFDL has advanced rapidly~\cite{xu2024fakeshield,yu2024diffforensics}; however, mainstream methods~\cite{ma2023iml, dong2022mvss, kwon2021cat, liu2022pscc, guo2023hierarchical} typically demonstrate strong performance only on specific datasets, limiting their generalizability. 
Recent methods~\cite{yu2024diffforensics, zhang2025training} attempt to leverage diffusion models for more precise identification of manipulation boundaries to alleviate this limitation.
Despite their advancements, these methods still exhibit considerable room for improvement in generalization.
Nowadays, MLLMs demonstrate impressive capabilities across a wide range of vision-language tasks, exhibiting substantial generalization potential~\cite{liu2023visual, zhu2023minigpt, tong2025medisee,openai_gpt5_blog_2025}.
Therefore, FakeShield~\cite{xu2024fakeshield} introduces the generalization capability of MLLMs into the image forgery detection domain through large-scale training.
Despite the success, it costs considerable training and computational resources.
Motivated by this observation, we raise a fundamental question: ``\textit{Do vanilla MLLMs inherently possess the generalization potential to address this problem, which has simply yet to be uncovered?}''

\begin{figure}[!t]
    \setlength{\belowcaptionskip}{-15pt}
    \centering
    \includegraphics[width=0.98\columnwidth]{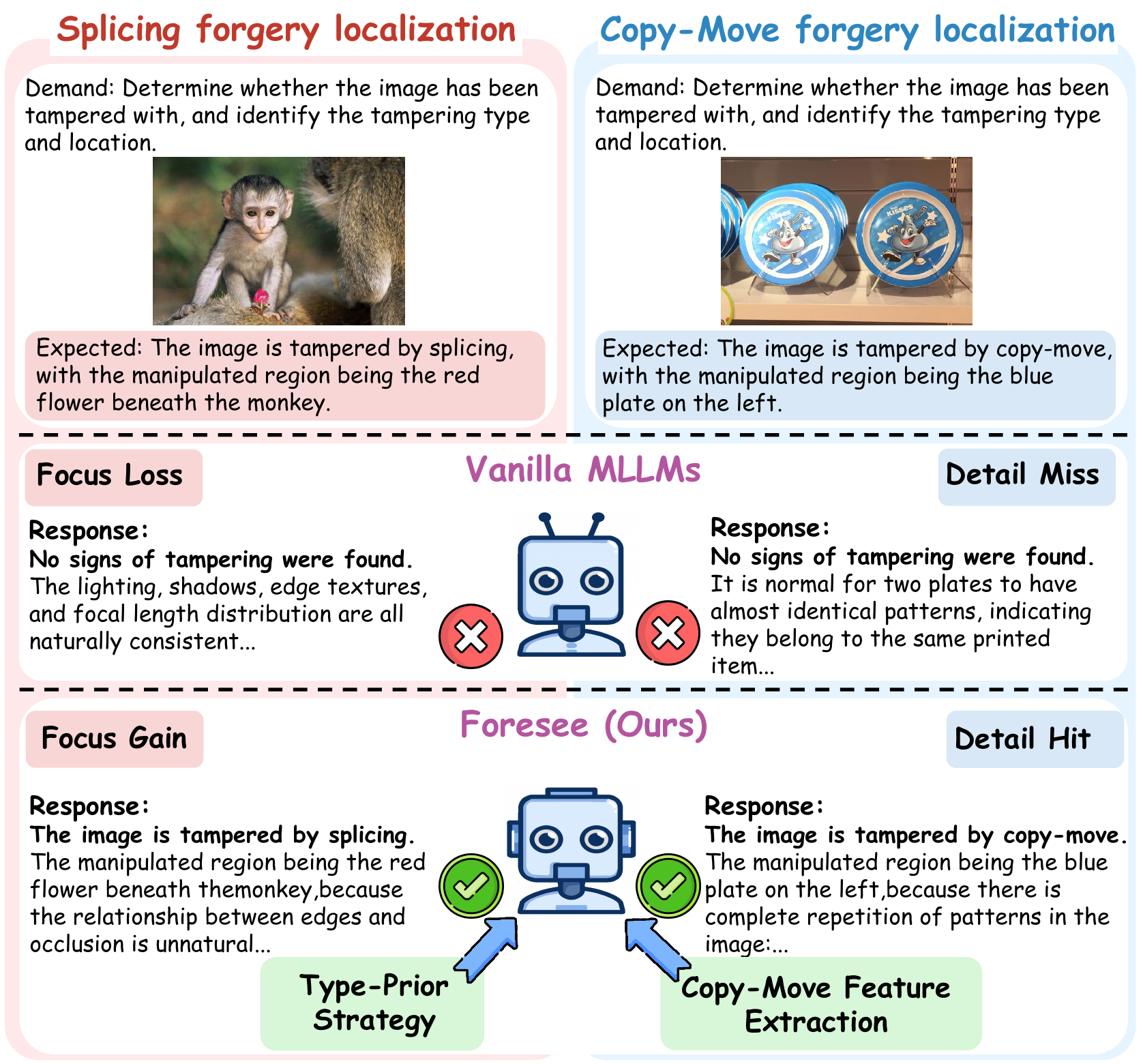}
    \caption{
Qualitative comparison between vanilla MLLMs and our Foresee pipeline on \textit{splicing} and \textit{copy-move} cases in the IFDL task, illustrating Foresee’s improved localization accuracy.
    }
    \label{fig:case1}
\end{figure}

Inspired by this, we first attempt to directly employ vanilla MLLMs~\cite{openai_gpt5_blog_2025, google_gemini_25_api_2025, Qwen3VL235B_2024, anthropic2024claude4} to address the image manipulation problem; however, the results are unsatisfactory, as shown in Figure~\ref{fig:case1}, their responses fall short of expectations—these models tend to overlook fine-grained tampering clues and often provide overly general or semantically plausible explanations instead of identifying the actual manipulations.
To overcome these limitations, we propose \textbf{Foresee}, a training-free MLLM-based pipeline tailored for interpretable image forgery analysis.
As illustrated in Figure~\ref{fig:teaser}, Foresee eliminates the need for additional training and enables a lightweight inference process, while surpassing existing MLLM-based methods in both tamper localization accuracy and the richness of textual explanations. 
The right side of Figure~\ref{fig:teaser} further demonstrates that the proposed pipeline achieves superior localization performance with significantly less training data compared to state-of-the-art methods, highlighting its strong generalization capability across diverse data distributions. 
Figure~\ref{fig:case1} intuitively illustrates performance that demonstrates the potential of vanilla MLLMs unleashed by Foresee in image forgery detection.
Our pipeline augments vanilla MLLMs with a type-prior-driven reasoning process and supplies copy-move-specific feature extraction hints, enabling accurate identification of various manipulation types (e.g., \textit{splicing}, \textit{copy-move}) and providing more insightful textual explanations. 

Specifically, Foresee leverages a type-prior-driven strategy to first determine the manipulation type of the input image, thereby guiding the MLLM to better focus on tampering traces relevant to each category. This step is followed by MLLM-guided inference and segmentation to jointly produce explanatory texts and localization masks. In addition, Foresee integrates a dedicated \textit{copy-move} detection module to compensate for the limited capability of vanilla MLLMs in recognizing self-consistent forgeries. Together, these components constitute a training-free, generalizable, and interpretable forensic pipeline for IFDL. Our main contributions are summarized as follows:
\begin{itemize}
    \item We present the first attempt to explore the direct application of vanilla MLLMs to the image forgery detection and localization (IFDL) task and propose a novel training-free pipeline named \textbf{Foresee}.
    Foresee not only alleviates generalization and interpretability limitations in existing IFDL approaches, but also substantially reduces dependence on large-scale training data and computationally resources that constrain interpretable models.
    \item We introduce a \textbf{F}lexible \textbf{F}eature \textbf{D}etector (\textbf{FFD}) module, which incorporates interchangeable training-free approaches for processing \textit{copy-move} manipulated images.
    \item Our work fills the gap in image forgery localization for training-free methods. Extensive experiments demonstrate that our approach achieves superior localization accuracy, provides more comprehensive textual explanations, and exhibits stronger generalization capability, outperforming existing IFDL methods across diverse tampering types including \textit{copy-move}, \textit{splicing}, \textit{removal}, \textit{local enhancement}, \textit{deepfake}, and \textit{AIGC-based editing}.
\end{itemize}

%% file: sec/relatedworks.tex
\section{Related Work}
\label{sec:related work}

\begin{figure*}[!t]
    \centering
    \vspace{-4pt}
    \includegraphics[width=1\textwidth]{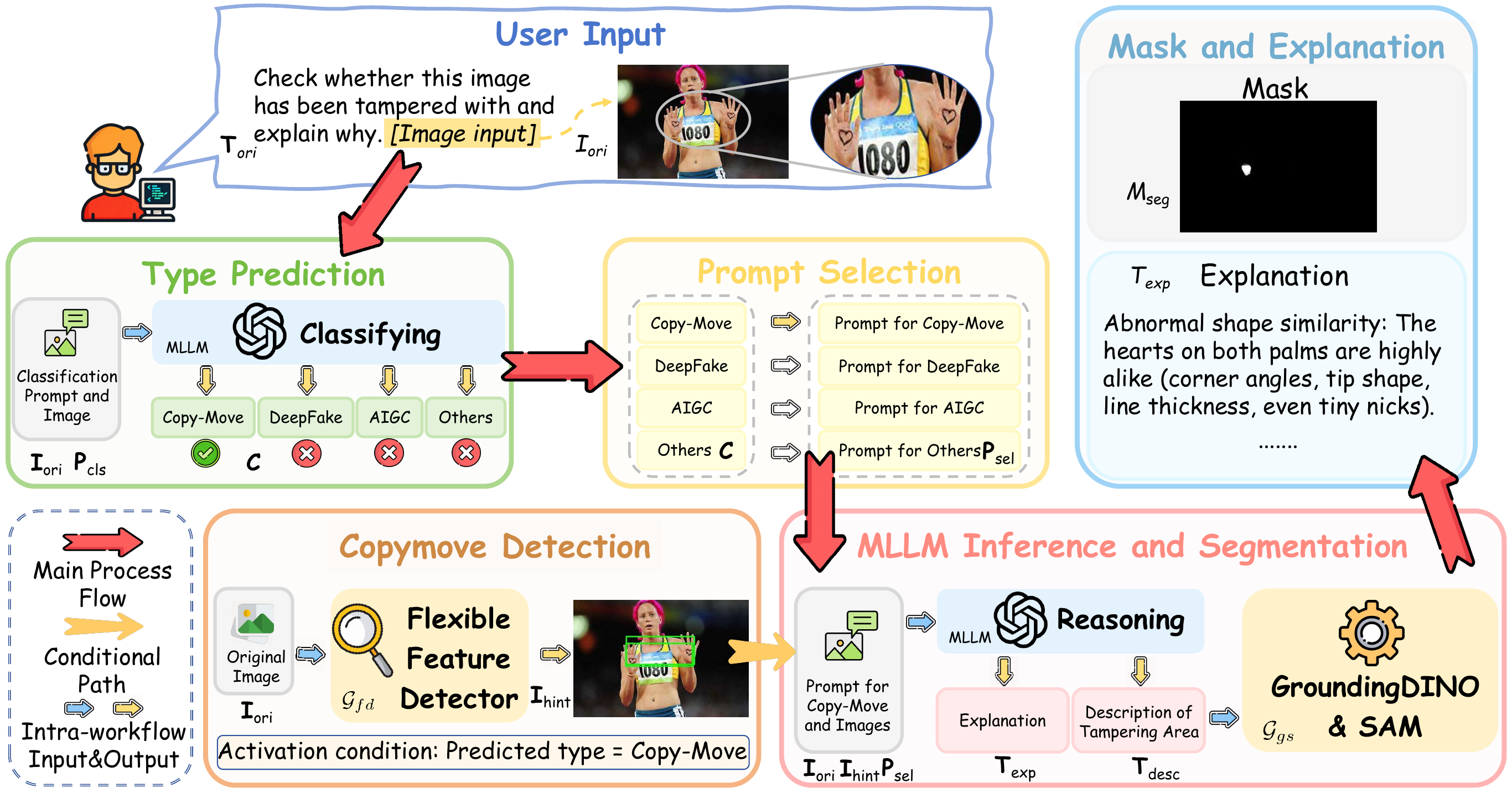}
    \caption{
    Overview of the proposed Foresee pipeline. 
    The input image $\mathbf{I}_{ori}$ and classification prompt $\mathbf{P}_{cls}$ are processed by $\operatorname{MLLM}$ to predict the classification label $\mathbf{C}$. If $\mathbf{C}$ indicates copy-move, the Flexible Feature Detector $\mathcal{G}_{dt}$ generates a hint image $\mathbf{I}_{hint}$. Then, $\mathbf{I}_{ori}$, $\mathbf{I}_{hint}$ (if available), and a task-specific prompt $\mathbf{P}_{sel}$ are fed into $\operatorname{MLLM}$ to produce a textual explanation $\mathbf{T}_{exp}$ and a concise description $\mathbf{T}_{desc}$ of the tampered region. Finally, $\mathbf{T}_{desc}$ guides the Grounded Segmentation Module $\mathcal{G}_{gs}$ to generate the tampering mask $\mathbf{M}_{seg}$.
    }
    \vspace{-4pt}
    \label{fig:framework}
\end{figure*}

\subsection{Image Forgery Detection and Localization}
Traditional forgery detection methods~\cite{salloum2018image, islam2020doa, li2018fast, zhu2018deep, li2019localization, wu2018busternet, fridrich2003detection,christlein2012evaluation,amerini2011sift} have primarily focused on identifying a single type of manipulation and generally lack robust localization capabilities. In recent years, mainstream approaches~\cite{kwon2021cat, ma2023iml, wu2019mantra, hu2020span, wu2022robust, guo2023hierarchical, dong2022mvss, liu2022pscc} for image forgery detection and localization have increasingly emphasized model generalization, aiming to detect a wide range of manipulation types. Most of these approaches are based on convolutional neural networks or transformer architectures, leveraging pixel-level or learned feature representations to identify forgery traces. 
Although these methods achieve considerable success, they still often fall short in terms of generalization to diverse manipulation types. Recently, more novel methods~\cite{yu2024diffforensics, wangforgdiffuser} employing diffusion models have emerged, which help to mitigate the generalization limitations of previous approaches. Concurrently, a fine-tuned large language model~\cite{xu2024fakeshield} for IFDL tasks not only improve generalization, but also introduce a new paradigm of textual interpretation for the task. However, these state-of-the-art methods typically require substantial computational resources, which presents additional challenges for practical deployment of image forgery detection and localization. Fortunately, a recent training-free method~\cite{zhang2025training}, utilizing diffusion and custom-designed modules aims to alleviate these constraints. Nevertheless, its localization performance still lags behind that of models trained specifically for this task. These limitations highlight the need for more efficient, generalizable, and interpretable approaches to image forgery detection and localization, motivating the development of our proposed method.

\subsection{MLLMs on Detection Tasks}
The application of Multimodal Large Language Models (MLLMs) to detection tasks has recently become an area of active research~\cite{liu2023visual,chen2023minigpt,tong2025medisee}. A prevalent approach involves fine-tuning MLLMs on high-quality image-text instruction datasets centered around detection scenarios. For instance, several recent methods~\cite{li2023blip, liu2023visual, zhu2023minigpt} construct detailed question–answer pairs that cover diverse detection cases. However, this process demands significant manual annotation and considerable computational resources, yet still struggles to generalize to unseen objects. Additionally, as most cutting-edge MLLMs remain closed-source~\cite{openai_gpt5_blog_2025, google_gemini_25_api_2025, anthropic2024claude4}, further instruction tuning is often impractical. In this work, we explore the inherent zero-shot detection capability of vanilla MLLMs to generalize across diverse types of image manipulations. Accordingly, we propose a novel training-free pipeline based on MLLMs, which decomposes the reasoning process through a Chain-of-Thought paradigm and leverages conventional forgery detection techniques as contextual cues for vanilla MLLMs. This strategy serves to fully unlock the potential of MLLMs in the domain of image forgery detection and localization.

%% file: sec/method.tex
\section{Methodology}
\label{sec:formatting}

While state-of-the-art MLLMs, such as GPT-5~\cite{openai_gpt5_blog_2025} and Gemini 2.5 Pro~\cite{google_gemini_25_api_2025}, exhibit impressive visual reasoning and recognition capabilities, their potential for identifying manipulated images is still unleashed. To fully unlock the capabilities of MLLMs in uncovering image forgeries, we introduce a training-free, MLLM-based pipeline named Foresee that decomposes the tampering detection process through the chain-of-thought~\cite{wei2022chain, kojima2022large} reasoning paradigm, enabling MLLMs to adapt more smoothly to IFDL tasks. The overview pipeline of the proposed Foresee is illustrated in Figure~\ref{fig:framework}, which comprises: forgery‑type prediction and task‑specific prompt selection (Section \ref{sec:Tamper_Type_Prediction}), followed by MLLM‑guided inference and segmentation to produce explanations and masks (Section \ref{sec:muti-modal-reasoning}). In addition, we introduce a dedicated \textit{copy-move} detection module to compensate for the limited abilities of vanilla MLLMs to recognize self-consistent forgeries (Section \ref{sec:FFD}). The pseudocode outlining the complete pipeline is provided in Algorithm \ref{alg:overall_pipeline}. 


\subsection{Forgery‑Type Prediction and Prompt Selection}
\label{sec:Tamper_Type_Prediction}
When directly applied to manipulated images, general-purpose MLLMs often struggle to maintain stable and coherent reasoning, primarily due to their tendency to conflate different types of tampering.
Therefore, we adopt a type-prior–driven strategy that guides the MLLM to perform tampering analysis in a step-by-step manner. Specifically, we first instruct the MLLM to identify the tampering type using a meticulously designed prompt that embeds descriptive features of different tampering categories, thereby facilitating more accurate classification. To facilitate this process, Foresee adopts a four-way taxonomy inspired by prior work~\cite{ma2023iml, nirkin2021deepfake}, covering \textit{copy-move}, \textit{deepfake}, \textit{AIGC}, and \textit{others}. These types correspond to distinct forensic evidence patterns: \textit{copy‑move} emphasizes intra‑image duplication and abnormally consistent textures~\cite{farid2009image, christlein2012evaluation, fridrich2003detection}; \textit{deepfake} focuses on facial appearance edits~\cite{tolosana2020deepfakes, verdoliva2020media}; \textit{AIGC-based} methods target inconsistencies between local regions and the whole image in illumination noise statistics~\cite{chai2020makes, lugmayr2022repaint}; the \textit{others} category covers splicing, removal, and local enhancement, which typically manifest as seam artifacts, boundary halos, and semantic inconsistencies~\cite{cozzolino2015splicebuster, li2017image, chen2008determining}.

\begin{table*}[h!]
\setlength{\abovecaptionskip}{2pt}
\centering
\caption{Detection performance comparison of various IFDL methods and vanilla MLLMs across multiple editing, deepfake, and AIGC-editing datasets, covering both traditional and generative manipulations types. The best and second-best results are highlighted in bold and underline, respectively. Explanation availability is indicated by \ding{51}/\ding{55}. \ssymbol{2}: Partially self-tested. \ssymbol{3}: Fully self-tested.}
\resizebox{\textwidth}{!}{%
\begin{tabular}{l|c|ccccc|c|c|c}
\toprule
\multirow{3}{*}{\parbox[c][4.5em][c]{2cm}{\raggedright \textbf{Method}}}
  & \multirow{3}{*}{\parbox[c][4.5em][c]{2cm}{\centering \textbf{Explanation}}}
  & \multicolumn{5}{c|}{\textbf{\makecell{Editing}}}
  & \multicolumn{1}{c|}{\textbf{DeepFake}}
  & \multicolumn{1}{c|}{\textbf{AIGC-Editing}}
  & \multirow{2}{*}{\textbf{Average}} \\
\cmidrule(lr){3-7} \cmidrule(lr){8-8} \cmidrule(lr){9-9}
& & CASIA1.0+ & Columbia & Coverage & NIST16 & IMD2020 & FaceApp & OpenForensics & \\ 
\cmidrule(lr){3-10}
& & \multicolumn{8}{c}{image-level AUC} \\ \midrule
 \rowcolor{purple!20}
 \multicolumn{10}{c}{\textit{\textbf{trained methods}}} \\
 HiFi-Net\ssymbol{2}~\cite{guo2023hierarchical} & \ding{55} & 46.0 & 68.0 & 34.0 & 58.8 & 62.0 & 56.0 & 28.0 & 50.4\\
 PSCC-Net\ssymbol{2}~\cite{liu2022pscc} & \ding{55} & \underline{90.0} & 78.0 & 84.0 & 61.3 & 67.0 & 48.0 & 39.5 & 66.8\\
 MVSS-Net\ssymbol{2}~\cite{dong2022mvss} & \ding{55} & 62.0 & 94.0 & 65.0 & \underline{72.3} & 75.0 & \underline{84.0} & 45.5 & 71.2\\ 
 FakeShield\ssymbol{2}~\cite{xu2024fakeshield} & \ding{51} & \textbf{95.0} & 98.0 & \textbf{97.0} & \textbf{74.8} & \textbf{83.0} & \textbf{93.0} & 37.5 & \textbf{82.5}\\
\rowcolor{green!20}
\multicolumn{10}{c}{\textit{\textbf{vanilla MLLMs}}} \\
GPT-5~\cite{openai_gpt5_blog_2025} & \ding{51} & 70.7 & 97.5 & 71.0 & 66.3 & 77.3 & 61.0 & 48.5 & 70.3\\ 
Gemini 2.5 Pro~\cite{google_gemini_25_api_2025} & \ding{51} & 72.4 & \underline{98.5} & 73.5 & 64.2 & 73.2 & 67.0 & 53.0 & 71.7\\
 
\rowcolor{blue!20}
 \multicolumn{10}{c}{\textit{\textbf{training-free methods}}} \\
Noise\ssymbol{3}~\cite{chen2008determining, fridrich2003detection, lukavs2006detecting} & \ding{55} & 28.1 & 46.3 & 69.5 & 17.3 & 21.9 & 6.0 & 3.5 & 27.5\\ 
Block\ssymbol{3}~\cite{amerini2011sift, fridrich2003detection} & \ding{55} & 25.8 & 40.6 & 76.0 & 12.4 & 23.7 & 2.5 & 5.0 & 26.6 \\ 
Point\ssymbol{3}~\cite{fridrich2003detection, christlein2012evaluation, wu2018busternet} & \ding{55} & 34.5 & 25.6 & 81.0 & 21.6 & 31.0 & 8.0 & 4.0 & 29.4 \\ 
Foresee(GPT-5) & \ding{51} & 83.2 & 97.5 & 85.5 & 69.0 & \underline{78.6} & \underline{84.0} & \underline{50.0} & 78.3 \\
\rowcolor{aliceblue!60}
Foresee(Gemini 2.5 Pro) & \ding{51} & 86.6 & \textbf{99.0} & \underline{87.5} & 68.3 & 74.1 & \textbf{93.0} & \textbf{57.5} & \underline{80.9} \\
\bottomrule
\end{tabular}
}
\label{table:image-auc}
\vspace{-4mm}
\end{table*}

As shown in Figure~\ref{fig:framework}, the input image $\mathbf{I}_{ori}$, together with the classification prompt $\mathbf{P}_{cls}$, is provided to the $\operatorname{MLLM}$ for joint processing, which outputs an image classification label $\mathbf{C}$. 
Notably, during this process, a tentative tampering type is assigned to every input image, regardless of whether manipulation is present; the following reasoning phase will determine whether manipulation exists. Subsequently, a corresponding type-aligned prompt $\mathbf{P}_{sel}$ is selected to steer the  $\operatorname{MLLM}$ toward the most discriminative tampering traces and spatial cues associated with that category.
The process can be formulated as:
\begin{equation}
    \mathbf{C} = \operatorname{MLLM}(\mathbf{I}_{ori},\mathbf{P}_{cls}),
\end{equation}
\begin{equation}
    \mathbf{P}_{\mathrm{sel}} = \mathbf{P}_{\mathrm{sel}}^{(C)}.
\end{equation}
This task-prior design incorporates the chain-of-thought paradigm, effectively alleviating the reasoning burden of the MLLM in subsequent inference stages.

\subsection{MLLM Inference and Segmentation}
\label{sec:muti-modal-reasoning}

Building upon the prompt selection and task-prior strategy described previously, this stage leverages the MLLMs to perform comprehensive tampering analysis, and utilizes its outputs to activate GroundingDINO~\cite{liu2024grounding} and SAM2~\cite{ravi2024sam} to accomplish mask segmentation. Specifically, the pipeline utilizes the input image $\mathbf{I}_{ori}$, a hint image cue tailored for \textit{copy-move} tampering $\mathbf{I}_{hint}$ (obtained by the \textbf{F}lexible \textbf{F}eature \textbf{D}etector module \(\mathcal{G}_{fd}\)), and the category-specific prompt $\mathbf{P}_{sel}$ to steer the MLLM’s reasoning process. The model generates a comprehensive, multi-faceted textual explanation $\mathbf{T}_{exp}$ and a concise yet effective description of the suspected tampering area $\mathbf{T}_{desc}$ (\textit{e.g., “a heart-shaped doodle on the athlete's left hand”}). The description $\mathbf{T}_{desc}$ is subsequently utilized by \textbf{Grounded Segmentation Module} \(\mathcal{G}_{gs}\), including GroundingDINO and SAM, where GroundingDINO leverages its strong text–region alignment capability to precisely localize the tampered area described by $\mathbf{T}_{desc}$, providing accurate spatial coordinates for subsequent segmentation. The grounded regions are subsequently utilized by SAM2, which produces precise segmentation masks \(\mathbf{M}_{\mathrm{seg}}\), thereby enhancing the accuracy of tampering localization. This process can be formulated as: 
\begin{equation}
\label{equa:I_hint}
    \mathbf{I}_{\mathrm{hint}} = \mathcal{G}_{fd}(\mathbf{I}_{ori}),
\end{equation}
\begin{equation}
    \mathbf{T}_{exp},\mathbf{T}_{desc} = \operatorname{MLLM}(\mathbf{I}_{ori},\mathbf{I}_{hint},\mathbf{P}_{sel}),
\end{equation}
\begin{equation}
    \mathbf{M}_{\mathrm{seg}} = \mathcal{G}_{gs}(\mathbf{T}_{desc}).
\end{equation}

\begin{algorithm}[H]
\caption{Overall Tampering Detection Pipeline}
\label{alg:overall_pipeline}
\begin{algorithmic}[1]
\State \textbf{Input:} User input image $\mathbf{I}_{ori}$, classification prompt $\mathbf{P}_{cls}$
\State \textbf{Output:} Tampering mask $\mathbf{M}_{seg}$, explanation $\mathbf{T}_{exp}$

\State Predict tampering type $\mathbf{C}$ = MLLM($\mathbf{I}_{ori}$,$\mathbf{P}_{cls}$)
\State Select category-specific prompt $\mathbf{P}_{sel}$ according to $\mathbf{C}$,
$\mathbf{P}_{sel}$ = $\mathbf{P}_{sel}^{(C)}$
\If{$\mathbf{C}$ == Copy-Move}
    \State Generate hint image $\mathbf{I}_{hint}$ = $\mathcal{G}_{fd}$($\mathbf{I}_{ori}$)
\Else
    \State  $\mathbf{I}_{hint} \gets \text{None}$
\EndIf
\State Input $\mathbf{I}_{ori}$, $\mathbf{I}_{hint}$ (if available), and $\mathbf{P}_{sel}$ to MLLM
\State Obtain explanation $\mathbf{T}_{exp}$ and description of tampering area $\mathbf{T}_{desc}$,
\Statex \hspace{\algorithmicindent} $\mathbf{T}_{exp}$,$\mathbf{T}_{desc}$  = MLLM($\mathbf{I}_{ori}$,$\mathbf{I}_{hint}$,$\mathbf{P}_{sel}$)
\State GroundingDINO localizes region based on $\mathbf{T}_{desc}$
\State SAM refines the region to generate final mask $\mathbf{M}_{seg}$,
\Statex \hspace{\algorithmicindent} $\mathbf{M}_{seg}$ = $\mathcal{G}_{gs}$($\mathbf{T}_{desc}$).
\State Return $\mathbf{M}_{seg}$ and $\mathbf{T}_{exp}$ as final results
\end{algorithmic}
\end{algorithm}

In Formula \ref{equa:I_hint}, \(\mathbf{I}_{hint}\) is produced by the Flexible Feature Detector module. This module is triggered only when the inferred tampering type is \textit{copy-move}, since we observe that MLLMs tend to struggle with \textit{copy-move} forgeries—primarily due to the subtlety of duplicated textures and the lack of well-defined tampering boundaries. We describe the design and implementation of this module in detail below.


\subsection{Flexible Feature Detector Module}
\label{sec:FFD}

To address the unique challenges posed by \textit{copy-move} tampering, where duplicated textures are often subtle and tampering boundaries are indistinct, we propose a \textbf{Flexible Feature Detector} module (\textbf{FFD}). 
This module is selectively activated when the image's predicted tampering type is \textit{copy-move}. It incorporates a set of interchangeable, training-free approaches, such as noise-based~\cite{chen2008determining, fridrich2003detection, lukavs2006detecting}, block-based~\cite{amerini2011sift, fridrich2003detection}, and keypoint-based methods~\cite{fridrich2003detection, christlein2012evaluation, wu2018busternet}, to extract and highlight potential regions of repeated textures within the original image $\mathbf{I}_{ori}$. They possess unique advantages over trained methods when dealing with \textit{copy-move} tampered images, such as being straightforward to implement, highly reproducible, and enabling fast inference, rendering them well-suited as hint image generators for the MLLM. Moreover, block-based and keypoint-based methods are specifically tailored for \textit{copy-move} tampering, effectively compensating for the inherent limitations of the MLLM in detecting duplicated textures. 
By leveraging these specialized feature extraction strategies, our pipeline enables the MLLM to discriminate self-consistent forgery patterns more effectively, thereby enhancing the performance in \textit{copy-move} manipulation localization.
The pseudocode of our proposed Foresee is presented in Algorithm~\ref{alg:overall_pipeline}.


%% file: sec/experiment.tex
\vspace{-1pt}
\section{Experiments}
\label{sec:experiment}

\begin{table*}[h!]
\setlength{\abovecaptionskip}{2pt}
\centering
\caption{Localization performance comparison of various IFDL methods and vanilla MLLMs across multiple editing, deepfake, and AIGC-editing datasets, covering both traditional and generative manipulations types. The best and second-best results are highlighted in bold and underline, respectively. Explanation availability is indicated by \ding{51}/\ding{55}. \ssymbol{2}: Partially self-tested. \ssymbol{3}: Fully self-tested.}
\resizebox{\textwidth}{!}{%
\begin{tabular}{l|c|ccccc|c|c|c}
\toprule
\multirow{3}{*}{\parbox[c][4.5em][c]{2cm}{\raggedright \textbf{Method}}}
  & \multirow{3}{*}{\parbox[c][4.5em][c]{2cm}{\centering \textbf{Explanation}}}
  & \multicolumn{5}{c|}{\textbf{\makecell{Editing}}}
  & \multicolumn{1}{c|}{\textbf{DeepFake}}
  & \multicolumn{1}{c|}{\textbf{AIGC-Editing}}
  & \multirow{2}{*}{\textbf{Average}} \\
\cmidrule(lr){3-7} \cmidrule(lr){8-8} \cmidrule(lr){9-9}
& & CASIA1.0+ & Columbia & Coverage & NIST16 & IMD2020 & FaceApp & OpenForensics & \\ 
\cmidrule(lr){3-10}
& & \multicolumn{8}{c}{pixel-level AUC} \\ \midrule
 \rowcolor{purple!20}
 \multicolumn{10}{c}{\textit{\textbf{trained methods}}} \\
 HiFi-Net\ssymbol{3}~\cite{guo2023hierarchical} & \ding{55} & 63.4 & 68.2 & 71.5 & 63.7 & 65.8 & 61.4 & 56.7 & 64.4\\
 PSCC-Net\ssymbol{3}~\cite{liu2022pscc} & \ding{55} & 76.9 & 80.6 & 74.1 & 69.4 & 77.2 & 67.4 & \textbf{68.9} & 73.5\\
 MVSS-Net\ssymbol{3}~\cite{dong2022mvss} & \ding{55} & 78.9 & 79.3 & 76.5 & 70.3 & 76.3 & 60.9 & 55.3 & 71.1\\ 
 FakeShield\ssymbol{3}~\cite{xu2024fakeshield} & \ding{51} & \textbf{82.9} & 83.1 & 69.6 & \underline{74.8} & \underline{80.6} & 66.2 & 67.6 & 75.0\\
\rowcolor{green!20}
\multicolumn{10}{c}{\textit{\textbf{vanilla MLLMs}}} \\
GPT-5~\cite{openai_gpt5_blog_2025} & \ding{51} & 69.8 & 79.6 & 60.5 & 71.8 & 74.6 & 61.7 & 56.2 & 67.7\\ 
Gemini 2.5 Pro~\cite{google_gemini_25_api_2025} & \ding{51} & 73.2 & 78.0 & 61.3 & 69.1 & 78.3 & 67.5 & 65.9 & 70.5\\
\rowcolor{blue!20}
 \multicolumn{10}{c}{\textit{\textbf{training-free methods}}} \\
Noise\ssymbol{3}~\cite{chen2008determining, fridrich2003detection, lukavs2006detecting} & \ding{55} & 56.1 & 53.8 & 59.4 & 53.9 & 52.0 & 52.9 & 54.1 & 54.6\\ 
Block\ssymbol{3}~\cite{amerini2011sift, fridrich2003detection} & \ding{55} & 62.5 & 58.7 & 58.2 & 52.8 & 57.7 & 49.5 & 40.9 & 54.3 \\ 
Point\ssymbol{3}~\cite{fridrich2003detection, christlein2012evaluation, wu2018busternet} & \ding{55} & 62.3 & 59.8 & 74.3 & 61.5 & 62.7 & 57.0 & 59.0 & 62.4 \\ 
Diffusion-LOC~\cite{zhang2025training} & \ding{55} & 58.7 & 68.2 & 62.2 & 55.6 & \raisebox{0.5ex}{\textbf{\rule{0.3cm}{0.6pt}}} & \raisebox{0.5ex}{\textbf{\rule{0.3cm}{0.6pt}}} & \raisebox{0.5ex}{\textbf{\rule{0.3cm}{0.6pt}}} & 61.2 \\ 
\rowcolor{aliceblue!60}
Foresee(GPT-5) & \ding{51} & 81.5 & \textbf{88.4} & \underline{80.2} & \textbf{83.5} & 78.5 & \underline{68.7} & 67.2 & \underline{78.3} \\
\rowcolor{aliceblue!60}
Foresee(Gemini 2.5 Pro) & \ding{51} & \underline{82.7} & \underline{86.7} & \textbf{81.9} & 72.2 & \textbf{85.2} & \textbf{72.1} & \underline{68.1} & \textbf{78.4} \\
\bottomrule
\end{tabular}
}
\label{table:pixel-auc}
\vspace{-4mm}
\end{table*}
\subsection{Experimental Setup}
\textbf{Dataset:} 
\label{Experimental Setup dataset}
Considering the availability and generality, we select several challenging benchmark datasets to evaluate our method. Among them, CASIA1+~\cite{dong2013casia}, Columbia~\cite{ng2009columbia}, IMD2020~\cite{novozamsky2020imd2020}, Coverage~\cite{wen2016coverage} and NIST16~\cite{guan2019mfc} are manipulated using traditional image editing tools; FaceApp~\cite{faceapp} serves as a deepfake tampering dataset; and OpenForensics~\cite{le2021openforensics} comprises samples with AIGC-based editing.
\par\vspace{0.3em}

\noindent\textbf{State-of-the-Art Methods:} To ensure a fair comparison, we select competitive methods that provide either open-source code or pre-trained models. To evaluate the \textbf{IFDL performance} of Foresee, we compare it against several SOTA trained methods as well as existing training-free approaches. The trained methods selected in our experiments include HiFi-Net~\cite{guo2023hierarchical}, PSCC-Net~\cite{liu2022pscc}, MVSS-Net~\cite{dong2022mvss} and FakeShield~\cite{xu2024fakeshield}. For consistency, all selected models are trained on the MMTD-Set~\cite{xu2024fakeshield} dataset, which encompasses manipulated images from various categories, including editing, deepfake, and AIGC-based editing. The training-free methods evaluated in our experiments include Diffusion-LOC~\cite{zhang2025training}, noise-based methods~\cite{chen2008determining, fridrich2003detection, lukavs2006detecting}, block-based methods~\cite{amerini2011sift, fridrich2003detection} and keypoint-based methods~\cite{fridrich2003detection, christlein2012evaluation, wu2018busternet}. Additionally, to assess the \textbf{textual output capability} of Foresee, we compare it with the interpretable forgery detection model FakeShield, as well as proprietary vanilla MLLMs, including GPT-5~\cite{openai_gpt5_blog_2025} and Gemini 2.5 Pro~\cite{google_gemini_25_api_2025}.
\par\vspace{0.3em}

\begin{table*}[h!]
\setlength{\abovecaptionskip}{2pt}

\centering
\caption{Localization performance comparison of various IFDL methods and vanilla MLLMs across multiple editing, deepfake, and AIGC-editing datasets, covering both traditional and generative manipulations types. The best and second-best results are highlighted in bold and underline, respectively. Explanation availability is indicated by \ding{51}/\ding{55}. \ssymbol{2}: Partially self-tested. \ssymbol{3}: Fully self-tested.}
\resizebox{\textwidth}{!}{%
\begin{tabular}{l|c|cccccccccc|cc|cc|cc}
\toprule
\multirow{3}{*}{\parbox[c][4.5em][c]{2cm}{\raggedright \textbf{Method}}}
  & \multirow{3}{*}{\parbox[c][4.5em][c]{2cm}{\centering \textbf{Explanation}}}
 & \multicolumn{10}{c|}{\textbf{\makecell{Editing}}}
  & \multicolumn{2}{c|}{\textbf{DeepFake}}
  & \multicolumn{2}{c|}{\textbf{AIGC-Editing}}
  & \multicolumn{2}{c}{\multirow{2}{*}{\textbf{Average}}}
  \\
  \cmidrule(lr){3-12} \cmidrule(lr){13-14} \cmidrule(lr){15-16}
  && \multicolumn{2}{c|}{CASIA1.0+}
  & \multicolumn{2}{c|}{Columbia}
  & \multicolumn{2}{c|}{Coverage}
  & \multicolumn{2}{c|}{NIST16}
  & \multicolumn{2}{c|}{IMD2020}
  & \multicolumn{2}{c|}{FaceApp}
  & \multicolumn{2}{c|}{OpenForensics}
  &&
  \\

\cmidrule(lr){3-18}
  & & IoU & F1 & IoU & F1 & IoU & F1 & IoU & F1 & IoU & F1 & IoU & F1 & IoU & F1 & IoU & F1 \\
\midrule
 \rowcolor{purple!20}
 \multicolumn{18}{c}{\textit{\textbf{trained methods}}} \\

 HiFi-Net\ssymbol{2}~\cite{guo2023hierarchical} & \ding{55} & 13.0 & 18.0 & 6.0 & 11.0 & 17.2 & 21.3 & 9.0 & 13.0 & 9.0 & 14.0 & 7.0 & 11.0 & 2.3 & 4.1 & 9.0 & 13.2\\
 PSCC-Net\ssymbol{2}~\cite{liu2022pscc} & \ding{55} & 36.0 & 46.0 & 64.0 & 74.0 & 17.6 & 22.4 & 18.0 & 26.0 & 22.0 & 32.0 & 12.0 & 18.0 & \textbf{9.4} & \textbf{15.8} & 25.6 & 33.5\\
 MVSS-Net\ssymbol{2}~\cite{dong2022mvss} & \ding{55} & 40.0 & 48.0 & 48.0 & 61.0 & 21.4 & 26.8 & 24.0 & 29.0 & 23.0 & 31.0 & 9.0 & 10.0 & 1.7 & 3.2 & 24.1 & 29.6\\ 
 FakeShield\ssymbol{2}~\cite{xu2024fakeshield} & \ding{51} & \textbf{54.0} & \textbf{60.0} & \underline{67.0} & \underline{75.0} & 16.5 & 19.7 & \underline{32.0} & \underline{37.0} & \textbf{50.0} & \textbf{57.0} & \underline{14.0} & \underline{22.0} & 5.9 & 8.2 & 34.2 & 39.8 \\
\rowcolor{green!20}
\multicolumn{18}{c}{\textit{\textbf{vanilla MLLMs}}} \\
GPT-5~\cite{openai_gpt5_blog_2025} & \ding{51} & 33.4 & 37.9 & 56.1 & 63.7 & 10.0 & 11.3 & 24.6 & 26.9 & 34.1 & 38.3 & 1.1 & 2.1 & 2.6 & 3.4 & 23.1 & 26.2\\ 
Gemini 2.5 Pro~\cite{google_gemini_25_api_2025} & \ding{51} & 35.2 & 39.3 & 55.2 & 61.8 & 10.6 & 12.1 & 21.4 & 23.9 & 38.7 & 43.5 & 10.3 & 15.7 & 9.7 & 15.0 & 25.9 & 30.2\\
\rowcolor{blue!20}
\multicolumn{18}{c}{\textit{\textbf{training-free methods}}} \\
Noise\ssymbol{3}~\cite{chen2008determining, fridrich2003detection, lukavs2006detecting} & \ding{55} & 0.4 & 0.7 & 0.2 & 0.3 & 12.1 & 18.1 & 3.4 & 4.7 & 1.9 & 3.2 & 0.1 & 0.1 & 0.1 & 0.1 & 2.6 & 3.9\\ 
Block\ssymbol{3}~\cite{amerini2011sift, fridrich2003detection} & \ding{55} & 17.0 & 23.8 & 15.4 & 25.2 & 19.8 & 24.7 & 3.8 & 5.2 & 3.5 & 5.8 & 0.1 & 0.1 & 1.5 & 2.8 & 8.7 & 12.5 \\ 
Point\ssymbol{3}~\cite{fridrich2003detection, christlein2012evaluation, wu2018busternet} & \ding{55} & 13.0 & 19.2 & 11.5 & 15.6 & 27.6 & 30.7 & 3.3 & 5.8 & 4.8 & 6.9 & 0.1 & 0.2 & 2.0 & 3.5 & 8.9 & 11.7 \\ 

\rowcolor{aliceblue!60}
Foresee(GPT-5) & \ding{51} & 43.3 & 52.1 & 65.2 & \textbf{77.8} & \underline{33.4} & \textbf{42.7} & \textbf{39.2} & \textbf{44.4} & 41.5 & 45.7 & 13.8 & 19.1 & 4.8 & 7.2 & \underline{34.5} & \underline{41.3} \\
\rowcolor{aliceblue!60}
Foresee(Gemini 2.5 Pro) & \ding{51} & \underline{44.1} & \underline{53.0} & \textbf{67.9} & 74.6 & \textbf{33.7} & \underline{42.1} & 30.3 & 34.5 & \underline{45.9} & \underline{50.9} & \textbf{21.1} & \textbf{28.3} & \underline{8.5} & \underline{11.6} & \textbf{35.9} & \textbf{42.1} \\
\bottomrule
\end{tabular}
}
\label{table:iouf1}
\vspace{-4mm}
\end{table*}

\noindent\textbf{Evaluation Metrics:} For forgery detection, we evaluate performance using image-level Area Under the ROC Curve (AUC). For forgery localization, we report pixel-level area under the curve (AUC), F1 scores, and Intersection over Union (IoU). For a fair and comprehensive evaluation of interpretability across all models, we employ GPT-5 as an automatic evaluator to assess the generated textual descriptions along four dimensions: (1) \textbf{Accuracy}, measuring how faithfully the description reflects the actual content and context of the manipulated regions; (2) \textbf{Level of Detail}, assessing the granularity and specificity of the described forgery artifacts; (3) \textbf{Hallucination}, identifying unsupported or irrelevant information not grounded in the image evidence; and (4) \textbf{Readability}, examining the fluency, coherence, and overall clarity of the generated text. For detection, a default threshold of $0.5$ is applied unless otherwise specified. As SAM2~\cite{ravi2024sam} employs a conservative segmentation strategy within bounding boxes, we lower its probability map threshold to $0.05$ to enhance localization accuracy.
\par\vspace{0.3em}
\noindent\textbf{Implementation Details:} Our method is \textbf{training-free} and does not require any additional fine-tuning or parameter optimization. For all experiments, we directly employ the proprietary models GPT-5~\cite{openai_gpt5_blog_2025} and Gemini 2.5 Pro~\cite{google_gemini_25_api_2025}. All inference is conducted using the official implementations on a single NVIDIA GeForce RTX 4090 GPU. Default settings are used throughout unless otherwise specified.

\subsection{Comparison with the State-of-the-Art Methods}

To ensure a comprehensive and fair evaluation of model performance in IFDL tasks, our experiments are designed to assess each method at both image-level and pixel-level. To examine the generalization capability of the models, we employ datasets from three distinct categories: editing, deepfake and AIGC-based editing. Furthermore, to evaluate the interpretability of model outputs, we develop a multidimensional assessment framework that scores textual explanations across four criteria.

\par\vspace{0.3em}
\noindent\textbf{Forgery detection evaluation:}
To demonstrate the superiority and generalization capability of our method for image forgery detection, we evaluate detection accuracy across three distinct categories of datasets: editing, deepfake and AIGC-based editing. As shown in Table~\ref{table:image-auc}, our Foresee achieves a significantly higher image-level AUC than other training-free methods, and attains the best performance on several individual datasets. Notably, in the evaluation of forgery detection, the average image-level AUC of Foresee is only marginally lower than that of the top trained methods, demonstrating the strong generalization capability of our approach. This also indicates that, even without any task-specific training, Foresee exhibits powerful intrinsic forgery detection ability.

\par\vspace{0.3em}
\noindent\textbf{Forgery localization evaluation:}
Tables~\ref{table:pixel-auc} and~\ref{table:iouf1} present the forgery localization results. Our method significantly outperforms other training-free methods and achieves leading pixel-level AUC, IoU, and F1 scores on multiple datasets. Notably, in terms of the average performance across these three metrics, the proposed training-free approach even surpasses the best trained method, highlighting its strong adaptability to diverse data distributions.

\begin{figure}[htbp]
    \centering
    \includegraphics[width=\columnwidth]{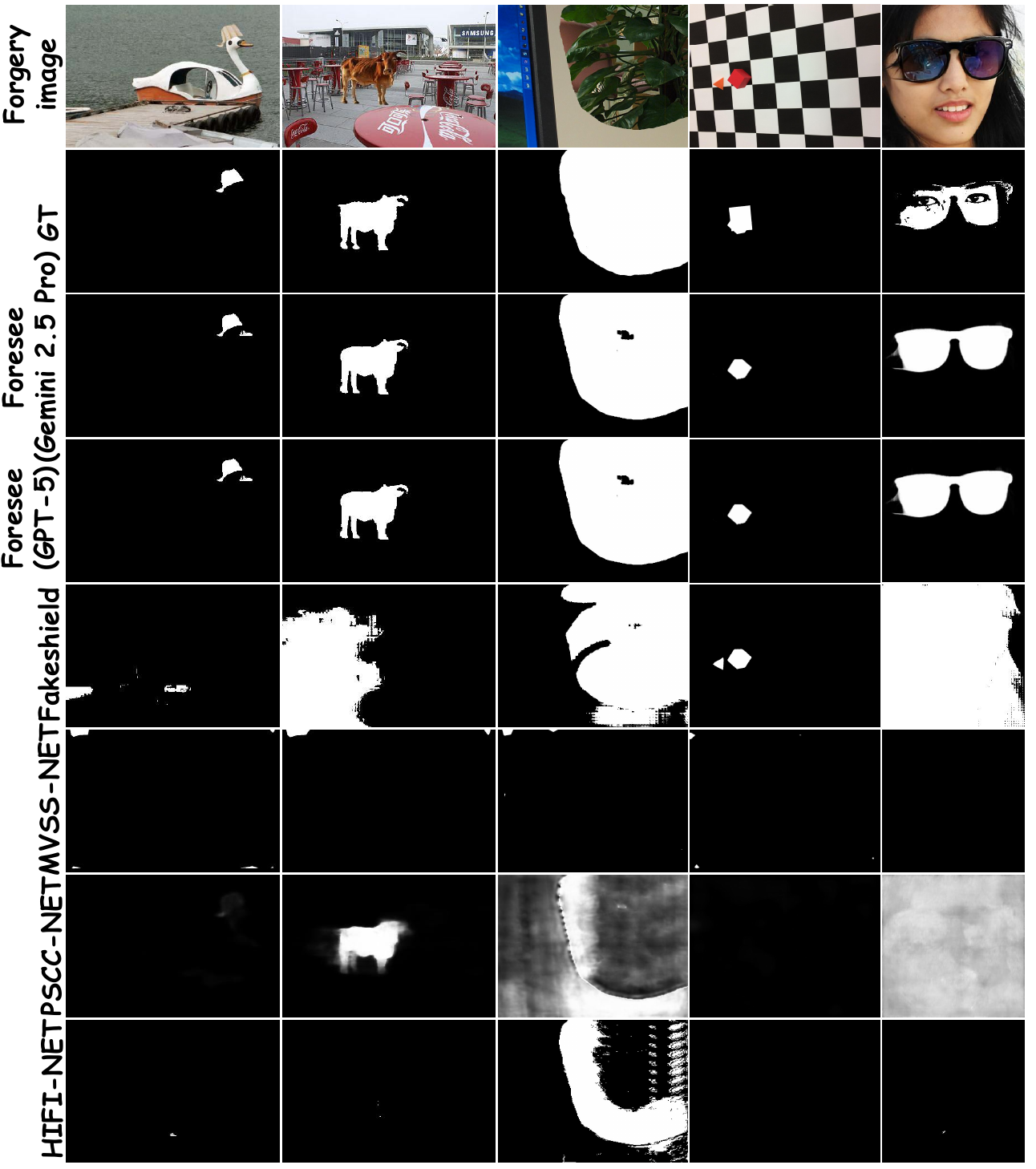}
    \caption{Visual comparison of predicted forgery masks from different methods on representative manipulated images. Ground truth masks and results from several networks are shown for qualitative assessment.} 
    \label{example}
    \vspace{-2em}
\end{figure}

Figure~\ref{example} illustrates qualitative comparisons of different forgery localization methods. Under the training-free paradigm, Foresee produces cleaner and more complete masks for manipulated regions across diverse data distributions, outperforming other networks in both coverage and precision. In particular, on face-related datasets, other models often focus solely on low-level forgery traces while failing to capture the semantic inconsistencies within facial regions, which limits their ability to provide coherent and accurate localization. These results further verify the generalization ability of Foresee in forgery localization.

\par\vspace{0.3em}
\noindent\textbf{Textual interpretability assessment:}
To quantitatively evaluate the quality of the explanatory text generated while localizing tampered regions by different methods, we borrow the powerful reasoning and comprehension capabilities of GPT-5~\cite{openai_gpt5_blog_2025}.
Specifically, we randomly sample 200 images from five editing type datasets. For each sample, the potentially manipulated image, the ground-truth image, and the generated textual explanation are jointly provided to GPT-5. GPT-5 then scores the textual explanation along four dimensions: accuracy, level of detail, hallucination, and readability, each scored on a five-point scale. This evaluation leverages the multimodal reasoning capabilities of MLLMs to provide a more holistic and human-aligned assessment compared to traditional automatic metrics such as cosine semantic similarity.
\begin{table}[h]
\vspace{-1mm}
\centering
\caption{Comparative results of vanilla MLLMs and explainable models in the image forgery detection and localization domain across different editing datasets. The best and second-best results are highlighted in bold and underline, respectively.}
\label{table:text}
\vspace{-2mm}
\scriptsize
\setlength{\tabcolsep}{2pt} 
\resizebox{\linewidth}{!}{
\begin{tabular}{l|cccc|c}
\toprule
\textbf{Method} & \textbf{Accuracy} & \textbf{Details} & \textbf{Hallucination} & \textbf{Readability} & \textbf{Average} \\
\midrule
\rowcolor{green!20}
\multicolumn{6}{c}{\textit{\textbf{vanilla MLLMs}}} \\
GPT-5~\cite{openai_gpt5_blog_2025}               & 1.26 & 4.10 & 2.22 & 4.94 & 3.13 \\
Gemini 2.5 Pro~\cite{google_gemini_25_api_2025}   & 0.89 & 4.25 & 1.40 & \textbf{5.00} & 2.89 \\
\rowcolor{blue!20}
\multicolumn{6}{c}{\textit{\textbf{explainable models for image forgery detection domain}}} \\
FakeShield~\cite{xu2024fakeshield}          & 2.41 & \underline{4.44} & 3.07 & \underline{4.95} & 3.72 \\
\rowcolor{aliceblue!60}
Foresee(GPT-5)      & \textbf{3.18} & 4.18 & \textbf{3.56} & 4.92 & \underline{3.91} \\
\rowcolor{aliceblue!60}
Foresee(Gemini 2.5 Pro)    & \underline{3.07} & \textbf{4.60} & \textbf{3.26} & \textbf{5.00} & \textbf{3.98} \\
\bottomrule
\end{tabular}
}
\vspace{-3mm}
\end{table}
As shown in Table~\ref{table:text}, our method achieves the highest scores across all four dimensions. These results indicate that the textual explanations generated by our model not only describe the manipulated regions with high precision, but also provide rich and detailed context. Moreover, our outputs effectively suppress hallucinations, ensuring reliability, and exhibit superior fluency and readability.

\par\vspace{0.3em}
\noindent\textbf{Robustness study:}
To assess the robustness of our method against common transmission degradations, we evaluate Foresee on the CASIA1+~\cite{dong2013casia} and Columbia~\cite{ng2009columbia} datasets under four degradation types: JPEG compression (quality factors of 70 and 80) and Gaussian noise (variances of 5 and 10). Performance across these degradation settings is summarized in Table~\ref{table:degrad}. As shown, Foresee exhibits minimal performance degradation under Gaussian noise and JPEG compression, demonstrating strong robustness in forgery localization even when image details are partially lost.

\begin{table}[h]
\centering
\caption{Localization performance on CASIA1.0+ and Columbia datasets under different degradation conditions.}
\label{table:degrad}
\vspace{-2mm}
\scriptsize
\resizebox{\linewidth}{!}{
\setlength{\tabcolsep}{7.5pt}
\begin{tabular}{l|cccccc}
\toprule
\multirow{2}{*}{\textbf{Method}} & \multicolumn{3}{c}{\textbf{CASIA1.0+}} & \multicolumn{3}{c}{\textbf{Columbia}} \\
\cmidrule(lr){2-7}
& AUC & IoU & F1 & AUC & IoU & F1 \\
\midrule
JPEG 70 & 79.5 & 42.8 & 48.9 & 86.9 & 64.7 & 73.7 \\
JPEG 80 & 79.7 & 42.9 & 49.1 & 87.3 & 64.9 & 73.9 \\
Gaussian 5 & 78.9 & 42.0 & 48.5 & 85.9 & 64.0 & 72.7 \\
Gaussian 10 & 77.1 & 41.2 & 47.7 & 84.4 & 61.5 & 70.5 \\
\rowcolor{aliceblue!60}
Foresee (GPT-5) & 81.5 & 43.3 & 52.1 & 88.4 & 65.2 & 77.8 \\
\bottomrule
\end{tabular}
}
\vspace{-3mm}
\end{table}


\subsection{Ablation Study}
This section analyzes the effectiveness of the type-prior strategy and the Flexible Feature Detector (FFD) module within the Foresee pipeline. We also conduct an ablation study on the choice of vanilla MLLMs employed in our framework.

\begin{table}[h]
\vspace{-1mm}
\centering
\caption{Localization performance of Foresee with and without the type-prior strategy on multiple diverse datasets.}
\label{table:type-prior}
\vspace{-2mm}
\scriptsize
\setlength{\tabcolsep}{8pt}
\begin{tabular}{l|cccccc}
\toprule
\multirow{2}{*}{\vspace{-1mm}\textbf{Method}}
& \multicolumn{2}{c}{\textbf{CASIA1.0+}} & \multicolumn{2}{c}{\textbf{Coverage}} & \multicolumn{2}{c}{\textbf{NIST16}} \\\cmidrule(lr){2-7}
& IoU & F1 & IoU & F1 & IoU & F1\\
\midrule
\makecell{Foresee(GPT-5)\\w/o type-prior} & 34.5 & 39.6 & 12.0 & 13.6 & 23.2 & 25.3\\
\midrule
\rowcolor{aliceblue!60}
Foresee(GPT-5)   & 43.3 & 52.1 & 33.4 & 42.7 & 39.2 & 44.4\\
\bottomrule
\end{tabular}
\vspace{-3mm}
\end{table}

\par\vspace{0.3em}
\noindent\textbf{Ablation on type-prior strategy:}
As shown in Table~\ref{table:type-prior}, incorporating type prediction and corresponding prompt selection notably enhances the localization performance of Foresee across all datasets. This demonstrates that adapting the prompting process with prior knowledge of manipulation type is crucial for achieving accurate and consistent forgery localization.

\begin{table}[h]
\vspace{-1mm}
\centering
\caption{Localization performance of Foresee with and without FFD module on copy-move datasets.}
\label{table:ffd}
\vspace{-2mm}
\scriptsize
\setlength{\tabcolsep}{8pt}
\begin{tabular}{l|cccccc}
\toprule
\multirow{2}{*}{\vspace{-1mm}\textbf{Method}}
& \multicolumn{3}{c}{\textbf{CASIA1.0+(CM)}} & \multicolumn{3}{c}{\textbf{Coverage}}\\\cmidrule(lr){2-7}
&AUC & IoU & F1 & AUC & IoU & F1 \\
\midrule
\makecell{Foresee(GPT-5)\\w/o   FFD} & 60.2 & 11.6 & 16.0 & 67.7 & 23.2 & 28.9\\
\midrule
\rowcolor{aliceblue!60}
Foresee(GPT-5)   & 76.8 & 24.2 & 31.4 & 80.2 & 33.4 & 42.7\\
\bottomrule
\end{tabular}
\vspace{-3mm}
\end{table}
\par\vspace{0.3em}
\noindent\textbf{Ablation on Flexible Feature Detector (FFD):}
We perform ablation experiments on Coverage~\cite{wen2016coverage} and the \textit{copy-move} subsets of CASIA1+~\cite{dong2013casia} datasets to evaluate the effectiveness of our FFD module. As shown in Table~\ref{table:ffd}, integrating FFD substantially enhances the localization performance of Foresee on \textit{copy-move} datasets. These results suggest that the FFD module effectively compensates for the limitations of vanilla MLLMs in capturing repetitive textures and local structural similarities, thereby enhancing their ability to detect \textit{copy-move} forgeries.

\begin{table}[h]
\vspace{-1mm}
\centering
\caption{Localization performance of Foresee with different MLLMs on CASIA1.0+ and Columbia datasets}
\label{table:mllms}
\vspace{-2mm}
\scriptsize
\setlength{\tabcolsep}{6pt}
\begin{tabular}{l|cccccc}
\toprule
\multirow{2}{*}{\vspace{-1mm}\textbf{Method}}
& \multicolumn{3}{c}{\textbf{CASIA1.0+}} & \multicolumn{3}{c}{\textbf{Columbia}}\\\cmidrule(lr){2-7}
       &AUC & IoU & F1 & AUC & IoU & F1\\
\midrule
Foresee(Qwen3-VL) & 78.0 & 38.7 & 46.5 & 79.8 & 60.9 & 68.9\\
Foresee(Claude Sonnet 4) & 77.2 & 36.2 & 44.6 & 78.3 & 46.2 & 52.8\\
\rowcolor{aliceblue!60}
Foresee(GPT-5)   & 81.5 & 43.3 & 52.1 & 88.4 & 65.2 & 77.8\\
\rowcolor{aliceblue!60}
Foresee(Gemimi 2.5 Pro)   & 82.7 & 44.1 & 53.0 & 86.7 & 67.9 & 74.6\\

\bottomrule
\end{tabular}
\vspace{-3mm}
\end{table}
\par\vspace{0.3em}
\noindent\textbf{Choices of vanilla MLLMs:}
Ablation results in Table~\ref{table:mllms} show that the choice of vanilla MLLMs has a notable impact on localization performance. Models powered by GPT-5~\cite{openai_gpt5_blog_2025} and Gemini 2.5 Pro~\cite{google_gemini_25_api_2025} consistently achieve superior results on CASIA1+~\cite{dong2013casia} and Columbia~\cite{ng2009columbia} datasets compared with those based on Qwen3-VL~\cite{Qwen3VL235B_2024} and Claude Sonnet 4~\cite{anthropic2024claude4}, highlighting the importance of advanced multimodal reasoning capabilities for forgery localization.

%% file: sec/conclusion.tex
\section{Conclusion}
\label{sec:formatting}


In this work, we introduce Foresee, a novel training-free pipeline for image forgery detection and localization (IFDL) that leverages the generalization potential of vanilla multimodal large language models (MLLMs). Foresee achieves strong generalization across diverse manipulation types and produces superior textual interpretations under multiple evaluation criteria. 
By employing a type-prior-driven strategy and utilizing a Flexible Feature Detector (FFD) module to specifically handle copy-move manipulations, Foresee demonstrates the generalization potential of vanilla MLLMs for forensics domain. 
Extensive experiments demonstrate that our work represents a significant advancement toward fully training-free and generalizable IFDL solutions, establishing Foresee as a solid foundation for future research in interpretable and multimodal image forensics. 
Foresee provides new insights into the intrinsic generalization capabilities of MLLMs and highlights their potential for broader visual understanding tasks.